\pdfoutput=1
\documentclass{article}

\usepackage{spconf}
\usepackage{amsmath,amssymb,amsfonts,graphicx}
\usepackage{placeins}
\usepackage[utf8]{inputenc}
\usepackage[colorlinks=true,linkcolor=blue,citecolor=blue,urlcolor=blue]{hyperref}
\usepackage[all]{hypcap}

\title{SAEC: Scene-Aware Enhanced Edge-Cloud Collaborative Industrial Vision Inspection with Multimodal LLM}

\name{Yuhao Tian\textsuperscript{\rm 1}, Zheming Yang\textsuperscript{\rm1,2*} \thanks{\textsuperscript{*}Corresponding author.}}

\address{ \textsuperscript{\rm 1}Institute of Computing Technology, Chinese Academy of Sciences, Beijing, China \\ \textsuperscript{\rm 2}Institute of AI for Industries, Nanjing, China}

\usepackage{color}

\begin{document}
%\ninept
%
\maketitle
\begin{abstract}
Industrial vision inspection requires high accuracy under stringent resource constraints, yet existing approaches face a fundamental trade-off. Multimodal LLMs (MLLMs) deliver strong reasoning capabilities but incur prohibitive computational costs, while lightweight edge models often fail on complex cases. In this paper, we present SAEC, a scene-aware enhanced edge-cloud collaborative industrial vision inspection framework with MLLM. The framework is composed of three synergistic components: (1) Efficient MLLM Fine-Tuning for Complex Defect Inspection, (2) Lightweight Multiscale Scene-Complexity Estimation, and (3) Adaptive Edge-Cloud Scheduler. Together, these modules enable robust defect detection by tailoring multimodal reasoning to scene complexity and dynamically balancing computation between edge and cloud resources. Experimental results on MVTec AD and KSDD2 datasets demonstrate that SAEC attains 85.11\% and 82.72\% accuracy, surpassing Qwen by 22.1\% and 20.8\%, and LLaVA by 33.3\% and 31.6\%. It also reduces runtime by up to 22.4\% and cuts energy per correct decision by 40\%-74\%. The code is available at \href{https://github.com/YuHao-Tian/SAEC}{https://github.com/YuHao-Tian/SAEC}.
\end{abstract}


\section{Introduction}
\label{sec:intro}
\begin{figure}[t]
  \centering
   \includegraphics[width=1\columnwidth]{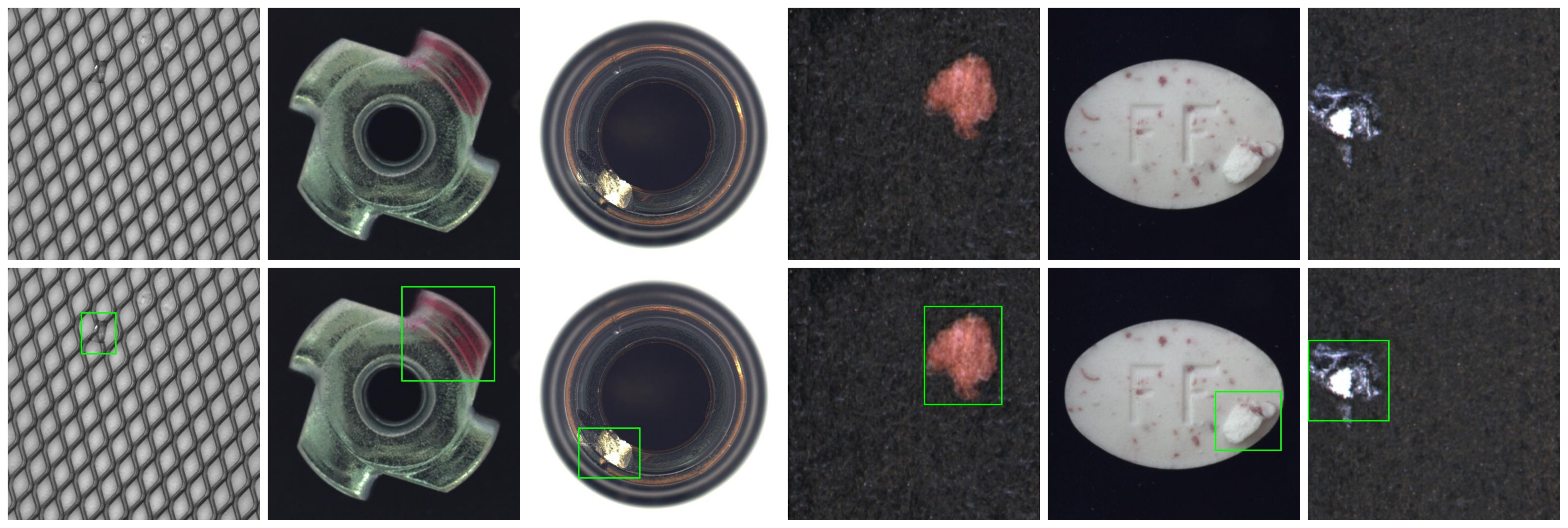}
  \caption{Examples of complex industrial vision inspection.}
  \label{figure1}
 % \vspace{-2mm}
\end{figure}

With the rapid development of artificial intelligence and the Industrial Internet of Things (IIoT) \cite{sisinni2018industrial111}, industrial vision inspection has emerged as a cornerstone technology in modern manufacturing \cite{malamas2003survey222}. Automated vision systems are increasingly deployed on assembly lines to detect surface defects \cite{zheng2021recent333}, structural inconsistencies, and assembly errors, thereby reducing reliance on human inspectors and minimizing production losses \cite{yang2021deep444}, as shown in Fig.~\ref{figure1}. Despite these advances, conventional vision inspection models remain largely constrained by their reliance on single-modality visual inputs \cite{yu2022efficient555}\cite{zheng2024md666}. As a result, their performance often degrades under complex and dynamic industrial conditions, such as varying illumination, background clutter, occlusions, or the coexistence of diverse materials \cite{conci2016integrated777}. These limitations lead to reduced robustness and generalization, making it difficult to meet the stringent demands of intelligent manufacturing for high accuracy, adaptability, and real-time responsiveness \cite{yang2023javp888}.

For these challenges, Multimodal Large Language Models (MLLMs) have recently attracted significant attention as a promising paradigm for industrial vision inspection \cite{zang2025contextual999}. By integrating heterogeneous modalities such as images and text, MLLMs exhibit enhanced semantic understanding and cross-modal reasoning capabilities \cite{wu2023multimodal101010}, which are crucial for accurately interpreting intricate industrial scenes. This capability enables them to identify subtle defect patterns that would otherwise be overlooked by unimodal vision systems \cite{mmcpf2024111111}. However, the adoption of MLLMs in real-world industrial environments is hindered by their massive parameter scale, high computational cost, and substantial memory footprint \cite{chen2024tomgpt-121212}. Deploying these models directly on resource-constrained edge devices often proves infeasible \cite{yang2024perllm131313}, while cloud-only processing introduces latency that disrupts the stringent real-time requirements of industrial pipelines \cite{yang2021intelligent14141414}.

To address these issues, we propose SAEC, a scene-aware enhanced edge-cloud collaborative industrial vision inspection framework with MLLM. The central idea of SAEC is to integrate scene-aware mechanisms with an edge-cloud collaborative architecture in order to balance accuracy and efficiency.  By harmonizing multimodal reasoning with adaptive task scheduling across edge and cloud resources, SAEC achieves robust defect detection under complex scenarios, while maintaining scalability, resource efficiency, and practicality for deployment in real-world manufacturing systems. Our contributions are as follows:
\begin{enumerate}
    \item We propose SAEC, the first framework that integrates scene-aware mechanisms with MLLM in an edge-cloud collaborative setting. It dynamically balances resource consumption between edge and cloud while adaptive inference based on the complexity of industrial inspection tasks.
    
    \item To enhance detection accuracy, we propose an efficient MLLM fine-tuning method suitable for complex defect detection.  It leverages parameter-efficient adaptation techniques combined with task-specific defect representations. 
    
    \item We present a comprehensive evaluation on the MVTec AD and KSDD2 datasets, demonstrating that SAEC achieves state-of-the-art accuracy while significantly reducing runtime and resource overhead compared to the MLLM baseline models.
\end{enumerate}

\section{Proposed Method}
\label{sec:method}

\begin{figure}[t]
  \centering
 \includegraphics[width=1\columnwidth]{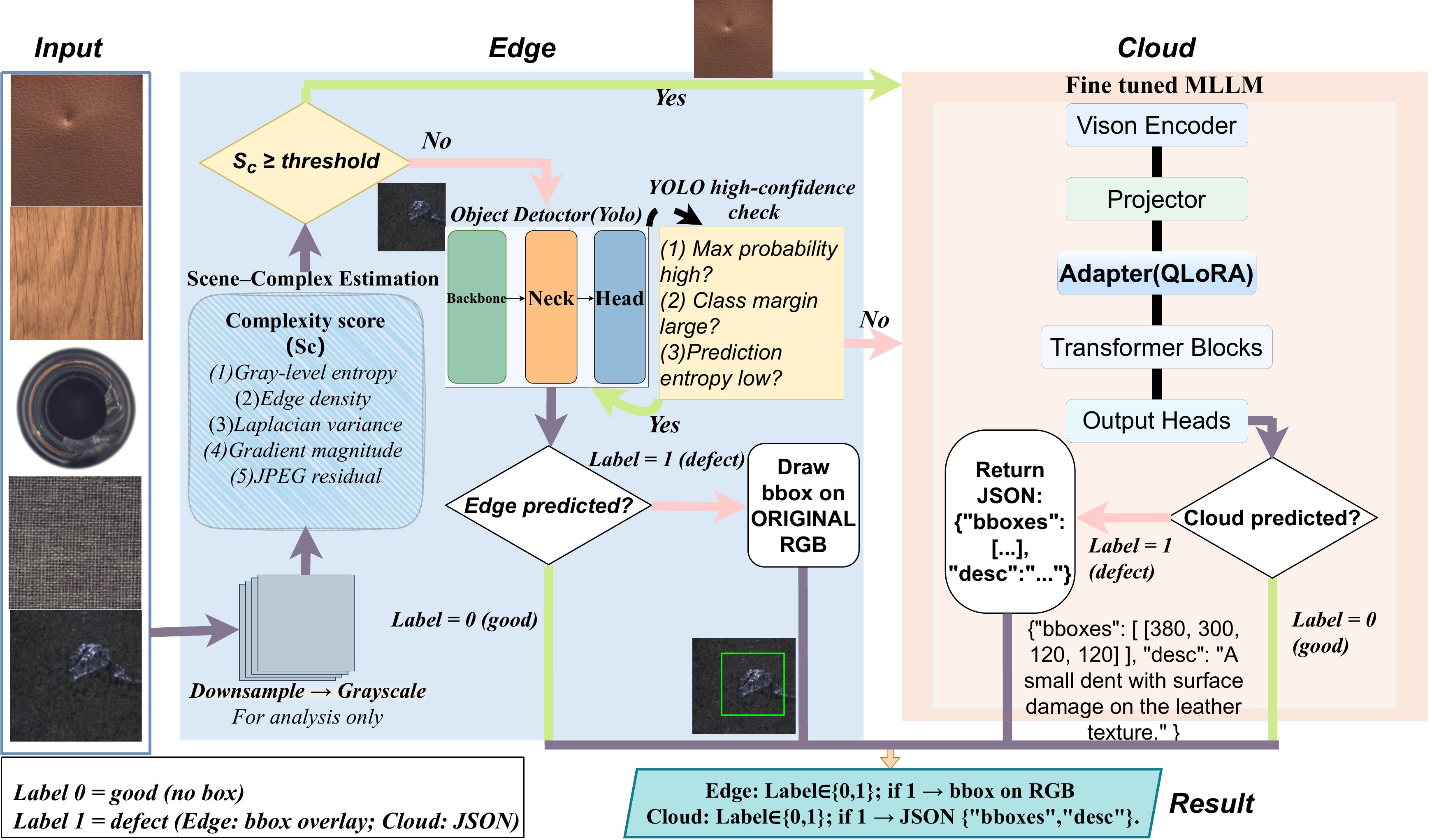}
  \caption{The overview of \textbf{SAEC}. A scene-complexity estimator drives budgeted routing: easy cases are finalized on the edge (lightweight object detector, such as the YOLO series), while complex or uncertain ones are escalated to the cloud (MLLM, such as Qwen and LLaVA series).}
  \label{fig:saec}
  \vspace{-2mm}
\end{figure}

We propose SAEC, a scene-aware enhanced edge-cloud collaborative 
industrial vision inspection framework with MLLM, as illustrated in Fig.~\ref{fig:saec}.  Its objective is to improve detection accuracy in complex defect inspection  (\emph{good}/\emph{defect})  while reducing wall-clock time and resource consumption. SAEC primarily consists of three parts: (1) \textit{Efficient MLLM Fine-Tuning for Complex Defect Inspection}, (2) \textit{Lightweight Multiscale Scene-Complexity Estimation}, and (3) \textit{Adaptive Edge-Cloud Scheduler}.

\subsection{Efficient MLLM Fine-Tuning for Complex Defect Inspection}
\label{subsec:qlora}
\begin{figure}[!t]
  \centering
  \includegraphics[width=1\columnwidth]{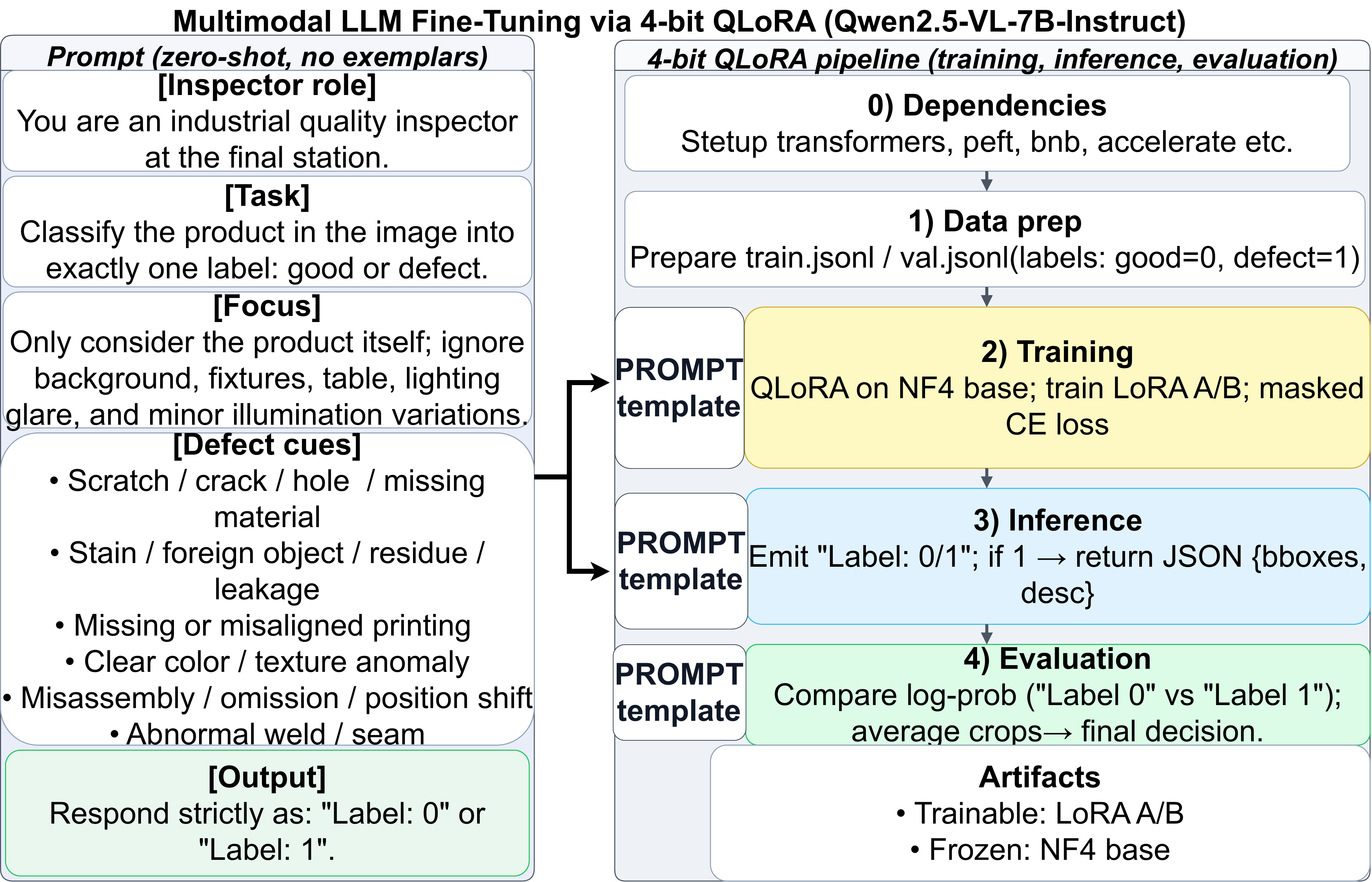}
  \vspace{-2mm}
  \caption{The overview of the efficient MLLM fine-tuning for complex defect inspection. \textbf{Left:} Zero-shot prompt used for both training and inference. \textbf{Right:} Pipeline with (0) dependencies, (1) data preparation, (2) QLoRA on NF4 base, (3) inference via last-token logits + threshold, and (4) evaluation by log-prob / crop averaging.}
  \label{fig:qlora}
  \vspace{-1mm}
\end{figure}

To enhance the accuracy of complex industrial inspections, we employ the 4-bit QLoRA \cite{dettmers2023qlora151515} to fine-tune the base model on dataset, which combines NormalFloat4 (NF4) quantization with low-rank adaptation, as shown in Fig.~\ref{fig:qlora}. Base model weights \(w\) are quantized block-wise into groups \(g\):
\begin{equation}
\begin{aligned}
\hat w &= \frac{w-\mu_g}{\sigma_g},\quad
k^\star = \arg\min_{k\in\mathcal{C}}\!\bigl|\hat w-c_k\bigr|,\quad
\tilde w = \sigma_g\,c_{k^\star}+\mu_g,
\end{aligned}
\end{equation}
where $\mathcal{C}$ is the 16-level NF4 codebook. The forward pass uses $\mathrm{DQ}(\mathrm{Q}(W))$ with bf16 compute, preserving model capacity while reducing memory and bandwidth requirements.

\textbf{Low-Rank Adaptation.} On selected attention and MLP projections $W_\ell$, we inject LoRA layers and train only the adapter parameters \cite{wei2024demonstrative161616}:
\begin{equation}
\begin{aligned}
W_\ell^\star=\mathrm{DQ}(\mathrm{Q}(W_\ell))+\Delta W_\ell,\qquad
\Delta W_\ell=(\alpha/r)\,A_\ell B_\ell,
\end{aligned}
\end{equation}
with $A_\ell\!\in\!\mathbb{R}^{d_{\text{out}}\times r}$ and $B_\ell\!\in\!\mathbb{R}^{r\times d_{\text{in}}}$, the overhead per layer is only $r(d_{\text{out}}+d_{\text{in}})$ parameters, enabling fine-tuning of a 7B model on a single GPU.

\textbf{Training and Inference.} Each sample is a short multimodal chat (image + concise instruction $\rightarrow$ strict answer). Let tokens $\mathbf{x}=[x_1,\ldots,x_T]$ interleave vision and text, and let $m_t\!\in\!\{0,1\}$ mask non-answer tokens so only the assistant’s answer contributes to the loss:
\begin{equation}
\mathcal{L}_{\mathrm{cls}}
= -\sum_{t=1}^{T} m_t \log p\!\big(x_t \mid x_{<t};\, \Theta, \{A_\ell,B_\ell\}\big),
\end{equation}
where $\Theta$ denotes the frozen 4-bit base model. In inference, we do not generate free text for screening. Instead, we read the last-token logits \((\ell_0,\ell_1)\), compute \(p_1=\mathrm{softmax}(\ell_0,\ell_1)_1\), and predict \emph{defect} if \(p_1\!\ge\!\tau\).

\textbf{Structured Defect Outputs.} If the predicted label is \texttt{1} (\emph{defect}), cloud decoding can continue under a constrained schema to emit normalized bounding box coordinates (JSON) and a one-sentence description:
\begin{equation}
\texttt{\{ "bboxes":[(x,y,w,h),\dots],\; "desc":"\dots" \}}.
\end{equation}
On the edge, if a sample is detected as a \emph{defect}, its bounding box is returned. These outputs are provided only for defect cases and do not influence the screening metrics.

\subsection{Lightweight Multiscale Scene-Complexity Estimation}
\label{subsec:complexity}

Scene complexity analysis is crucial for efficient edge-cloud collaborative inference \cite{yang2023javp888}. To effectively distinguish the difficulty of sample detection, we propose a lightweight multiscale scene-complexity estimation model as follows:

\begin{equation}
\label{eq:Sc}
S_c = w_1 H_I + w_2 E_d + w_3 \frac{\log(1+\sigma_L^2)}{8} + w_4 \frac{\bar{M}_S}{16} + w_5 r_J,
\end{equation}
where $\mathbf{w}=[w_1,\dots,w_5]$ are weight factors. The denominators 8 and 16 are used for feature scale alignment, eliminating the need for separate normalization. The calculation method for core metrics is as follows:

 (1) \textbf{Intensity Entropy ($H_I$):} Calculated from the normalized 256-bin histogram ($p_k$) of the grayscale image $G$. The formula is $H_I = -\frac{1}{\ln 256}\sum_{k=0}^{255}p_{k}\ln(p_{k}+\epsilon)$, with $\epsilon=10^{-12}$ for numerical stability.
 
 (2) \textbf{Edge Density ($E_d$):} Defined as the proportion of edge pixels in a binary edge map $E$ obtained using a Canny detector. The density is calculated as $E_d = \frac{1}{c^2}\sum_{u,v}\mathbf{1}[E(u,v)>0]$, where $c=192$. 
 
 (3) \textbf{Laplacian Variance ($\sigma_L^2$):} The variance of the Laplacian of the image, $\sigma_L^2 = \text{Var}(\nabla^2 G)$. To compress its dynamic range, the log-transformed value $\log(1+\sigma_L^2)$ is used in the final score. 
 
 (4) \textbf{Mean Sobel Magnitude ($\bar{M}_S$):} The average magnitude of the image gradient, computed using a $3{\times}3$ Sobel operator. The magnitude is $M_S=\sqrt{G_x^2+G_y^2}$, and its mean is $\bar{M}_S=\frac{1}{c^2}\sum_{u,v}M_S(u,v)$. 
 
 (5) \textbf{JPEG Residual ($r_J$):} The mean absolute difference between the original grayscale image $I$ and its decompressed version $I'$ after a JPEG compression-decompression cycle. It is computed as $r_J=\frac{1}{255}\,\text{mean}(|I-I'|)$.

\subsection{Adaptive Edge-Cloud Scheduler}
\label{subsec:selector}

The adaptive edge-cloud scheduler is designed to balance efficiency and accuracy through calibrated thresholding and dynamic routing. A small held-out split is used for temperature scaling to calibrate both prediction heads: the complexity percentile $\tau_S=\mathrm{Perc}_{1-\rho}(S_c)$ determined by the budget $\rho$, the edge acceptance thresholds $(\tau_s, \tau_m, \tau_h)$, and the cloud confidence threshold $\tau$.

At runtime, inputs with $S_c \ge \tau_S$ are routed to the cloud, while those with $S_c < \tau_S$ are processed on the edge only if they satisfy the confidence conditions $(s_{\max}\ge\tau_s)\wedge(m\ge\tau_m)\wedge(H_p\le\tau_h)$. Otherwise, they are also sent to the cloud. This mechanism ensures that only reliable predictions are handled locally, while complex cases are escalated. The scheduler executes complexity scoring in batches with negligible overhead, while edge and cloud processing branches operate in parallel. The overall latency is modeled as:
\begin{equation}
T_{\text{total}} = T_{\text{cpx}} + \max\!\big(T_{\text{edge}},\,T_{\text{cloud}}\big).
\end{equation}
This formulation ensures a consistent accounting of runtime and energy costs across different methods, enabling fair and robust evaluation of scheduling efficiency.

\section{Experiments}
\label{sec:experiments}

\subsection{Experimental Setup}
\label{subsec:setup}
All experiments are conducted on a single NVIDIA A100 (40\,GB) GPU (cloud) and an 8-core Intel Xeon Platinum 8575C CPU (edge). YOLO-11s is deployed at the edge, and Qwen-2.5-VL-7B is deployed in the cloud. We evaluate on two datasets: \emph{MVTec AD}~\cite{bergmann2019mvtec} and \emph{KolektorSDD2 (KSDD2)}~\cite{Bozic2021COMIND}. Both are converted to a binary format \texttt{val/\{good, defect\}} with a unified directory structure and an approximately 1:1 class ratio. Our baselines include YOLO-11s, Qwen-2.5-VL-7B\cite{qwen2024qwen25vl181818}, and LLaVA-1.5-7B \cite{liu2023visual171717}. We set the scene-aware weight factors as: \(\mathbf{w}=[0.30,\,0.25,\,0.20,\,0.15,\,0.10]\). Evaluation metrics  include accuracy, runtime \(T_{\text{total}}\), and resource overhead. All methods use identical precision, input resolution, and batching for fair comparison. 

\subsection{Main Results}
\label{subsec:main_result}

\textbf{Accuracy Comparison.}
As shown in Table~\ref{tab:acc_main}, SAEC achieves the highest accuracy on both datasets, substantially outperforming both YOLO-11s and multimodal LLM baselines. On the MVTec AD dataset, SAEC surpasses Qwen by 22.1\% and LLaVA by 33.3\%, while on KSDD2, the improvements are 20.8\% and 31.6\%, respectively. These margins indicate that the proposed framework is not only superior to lightweight edge models but also consistently outperforms state-of-the-art multimodal LLMs.

\begin{table}[h]
\centering
\caption{Accuracy (\%) comparison results. Higher is better.}
\label{tab:acc_main}
\resizebox{\linewidth}{!}{
\begin{tabular}{|l|c|c|c|c|}
\hline
\textbf{Dataset} & \textbf{YOLO-11s} & \textbf{Qwen-2.5-VL-7B} & \textbf{LLaVA-1.5-7B} & \textbf{SAEC (Ours)} \\ \hline
MVTec AD & 51.47 & 63.04 & 51.80 & \textbf{85.11} \\ \hline
KSDD2    & 51.32 & 61.94 & 51.12 & \textbf{82.72} \\ \hline
\end{tabular}
}
\end{table}

\textbf{Runtime Comparison.}
Our proposed SAEC framework achieves the fastest runtime, as detailed in Table~\ref{tab:time_main}. On MVTec AD, SAEC's total inference time is 172.8\,s, which is 22.4\% faster than Qwen (222.7\,s) and 13.7\% faster than LLaVA (200.3\,s). On KSDD2, SAEC takes 154.2\,s, which is 11.2\% faster than Qwen (173.6\,s) and 1.8\% faster than LLaVA (157.0\,s). This demonstrates that SAEC can effectively reduce cloud load and enhance inference efficiency through its scene-aware adaptive edge-cloud scheduler.

\begin{table}[h]
\centering
\caption{Runtime comparison results. Lower is better.}
\label{tab:time_main}
\resizebox{\linewidth}{!}{
\begin{tabular}{|l|c|c|c|}
\hline
\multicolumn{1}{|c|}{\textbf{Metric}} & \textbf{Qwen-2.5-VL-7B} & \textbf{LLaVA-1.5-7B} & \textbf{SAEC (Ours)} \\ \hline
\multicolumn{4}{|c|}{\textit{MVTec AD}} \\ \hline
Total time (s) & 222.7 & 200.3 & \textbf{172.8} \\ \hline
Avg time per image (s)   & 0.230 & 0.207 & \textbf{0.179} \\ \hline
\multicolumn{4}{|c|}{\textit{KSDD2}} \\ \hline
Total time (s) & 173.6 & 157.0 & \textbf{154.2} \\ \hline
Avg time per image (s)   & 0.244 & 0.221 & \textbf{0.217} \\ \hline
\end{tabular}
}
\end{table}

\begin{figure}[th]
\centering
 \includegraphics[width=1\columnwidth]{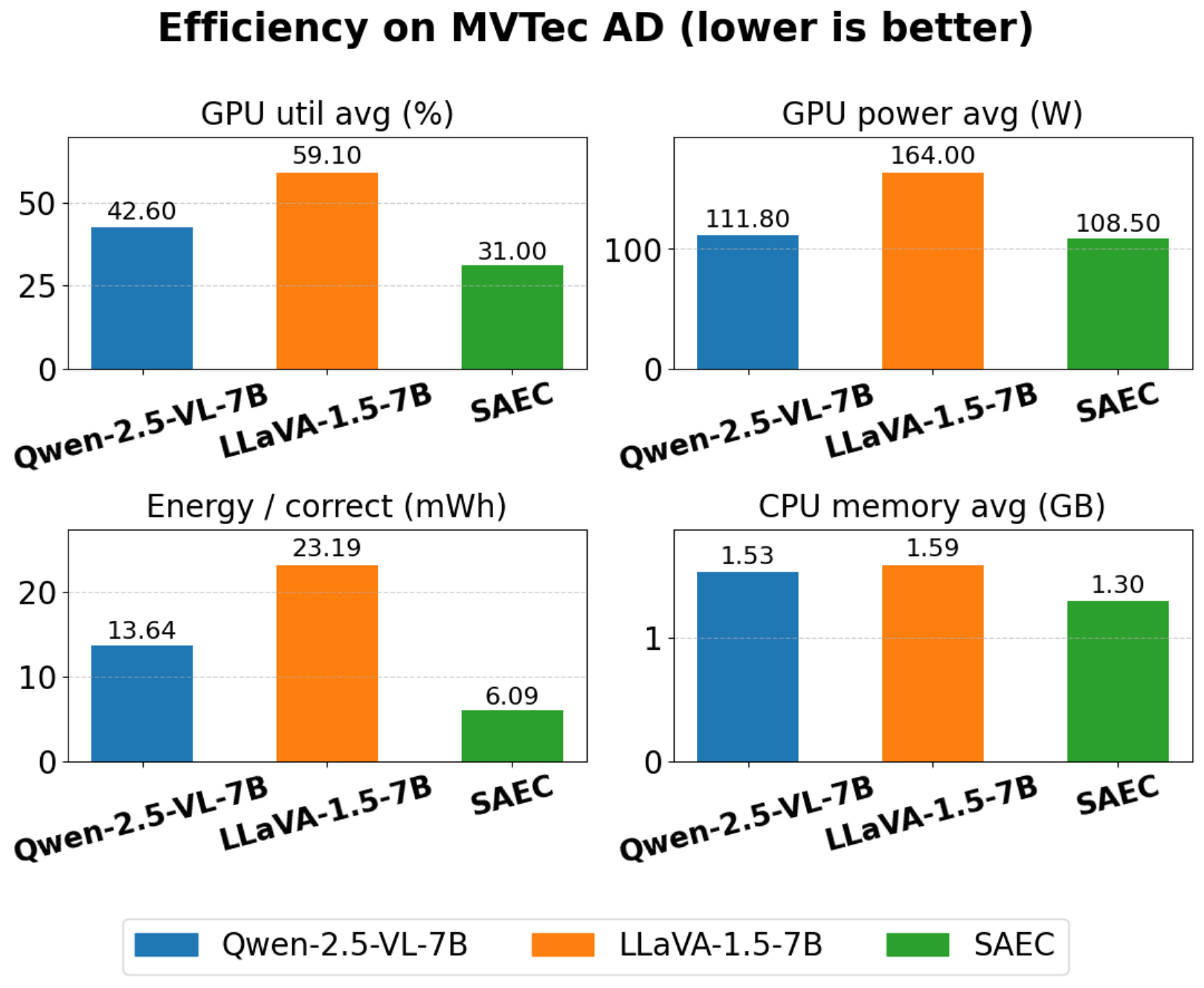}
\caption{Resource efficiency comparison on MVTec AD. Lower is better for all metrics.}
\label{fig:eff_mvtec}
\end{figure}

\textbf{Resource Efficiency Comparison.} \textit{(1)MVTec AD:} As shown in Fig.~\ref{fig:eff_mvtec}, SAEC is the most resource-efficient method. SAEC lowers GPU utilization to 31.0\% (27.2\% and 47.5\% below Qwen and LLaVA) and average GPU power to 108.5\,W. Most notably, the energy per correct prediction drops to just 6.09\,mWh, a reduction of 55.3\% compared to Qwen and 73.7\% compared to LLaVA. CPU memory usage remains minimal at 1.30\,GB, confirming that offloading easy cases to the CPU is memory-efficient.

\begin{figure}[t]
\centering
\includegraphics[width=0.92\linewidth]{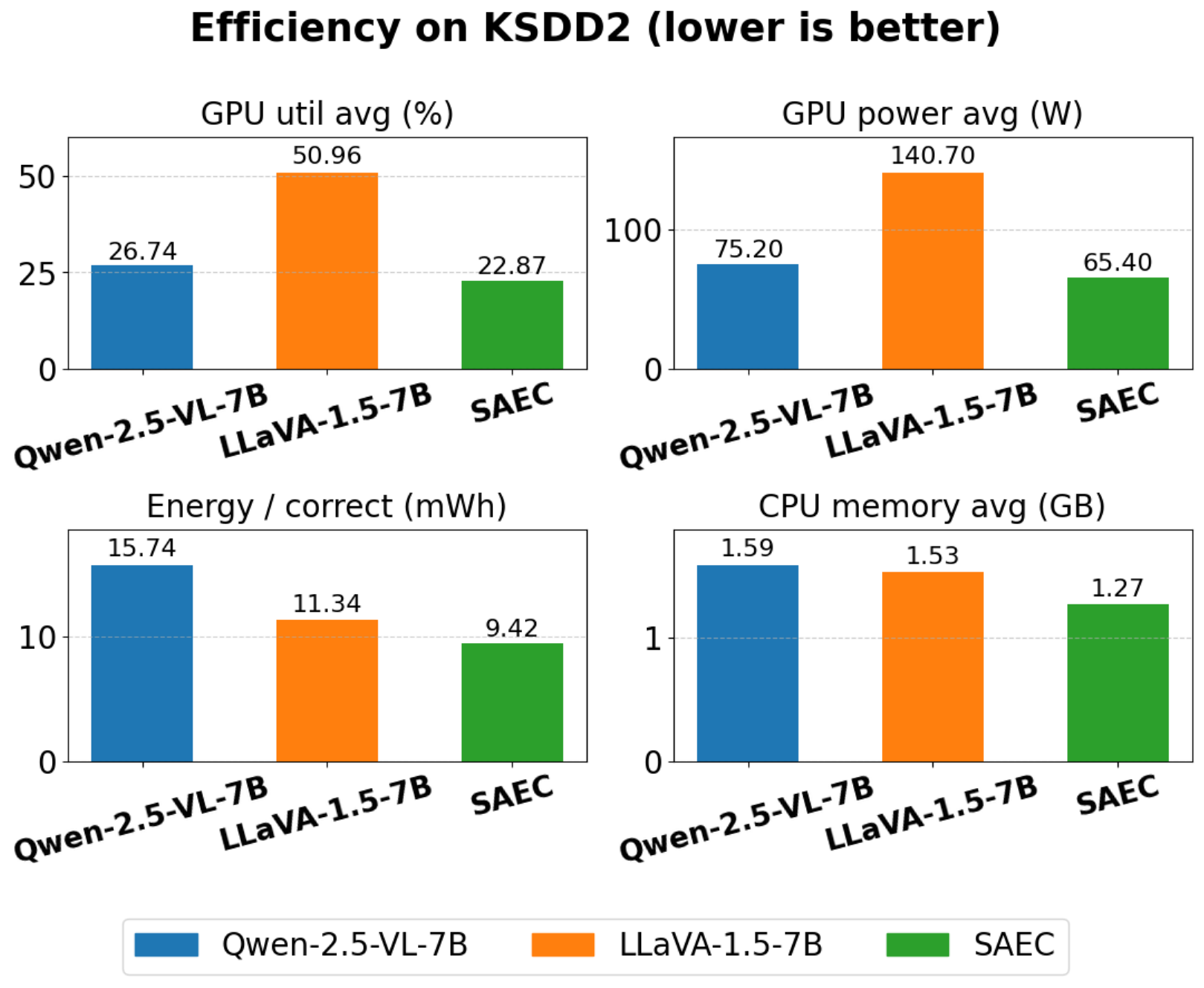}
\caption{Resource efficiency comparison on KSDD2. Lower is better for all metrics.}
\label{fig:eff_ksdd2}
\end{figure}

\textit{(2)KSDD2:} The efficiency gains are pronounced on KSDD2, as shown in Fig.~\ref{fig:eff_ksdd2}. SAEC reduces GPU utilization to 22.87\% (a 14.5\% relative reduction vs. Qwen and 55.1\% vs. LLaVA) and GPU power to 65.4\,W (13.0\% and 53.5\% lower, respectively). This culminates in an energy cost of only 9.42\,mWh per correct prediction (40.1\% and 16.9\% lower). The average CPU memory remains the lowest at 1.27\,GB. These results suggest that SAEC not only conserves GPU resources but also achieves a favorable energy–accuracy trade-off, making it particularly suitable for long-term industrial deployment under tight resource budgets.

\subsection{Ablation Study}
\textbf{Ablation Study I: Removing MLLM Fine-Tuning module:}
This removal causes a marked accuracy drop, \textbf{15.20\%} on \emph{MVTec AD} and \textbf{19.99\%} on \emph{KSDD2}.
In contrast, the computational overhead is negligible. Runtime remains within the same order (change $<2\%$), energy consumption is comparable, and CPU memory usage is virtually unchanged.
This shows MLLM fine-tuning is essential for the SAEC, delivering substantial accuracy gains at a minimal resource cost.

\textbf{Ablation Study II: Removing Scene-Aware Edge-Cloud Collaboration.} This removal significantly increases the computational burden: runtime rises by \textbf{11\%} and total energy consumption by \textbf{15\%}, accompanied by noticeably higher instantaneous GPU and CPU utilization. This shows that scene-aware edge–cloud collaboration can effectively reduce resource overhead and enhance inference efficiency.

\section{Conclusion}
In this work, we presented SAEC, a scene-aware enhanced edge–cloud collaborative framework for industrial vision inspection with MLLM. SAEC is designed to simultaneously improve detection accuracy while reducing runtime and resource overhead in complex industrial environments. On MVTec AD and KSDD2, SAEC significantly outperforms strong MLLM baselines, improving accuracy by up to 33.3\% while cutting runtime by up to 22.4\% and reducing energy per correct decision by up to 74\%. Overall, SAEC provides a resource-efficient and scene-adaptive solution for industrial vision inspection.

\bibliographystyle{IEEEbib}
\bibliography{refs}

\end{document}